\pdfoutput=1
\documentclass[10pt,twocolumn,letterpaper]{article}
\usepackage[dvipdfmx]{graphicx}
\usepackage{arxiv}
\usepackage{times}
\usepackage{epsfig}
\usepackage{amsmath}
\usepackage{amssymb}

\usepackage{bibunits}

\usepackage{times}
\usepackage{amsfonts}
\usepackage{bm}
\usepackage[dvipsnames]{xcolor}
\usepackage{multirow}
\usepackage{booktabs}

\usepackage[ruled,vlined]{algorithm2e}
\usepackage{setspace}
\usepackage{verbatim}
\usepackage{tabularx}

\newcommand{\citep}[1]{\cite{#1}}
\newcommand{\citet}[1]{\cite{#1}}
\newcommand{\clevr}{\textit{CLEVR}}
\newcommand{\task}{\textit{CLEVR Ask}}
\newcommand{\gw}{\textit{GuessWhat?!}}
\newcommand{\coco}{\textit{MSCOCO}}
\newcommand{\oracle}{\textit{Oracle}}
\newcommand{\questioner}{\textit{Questioner}}

\newcommand{\Simage}{$\mathcal{I}$}
\newcommand{\Sdialogue}{$\mathcal{D}$}
\newcommand{\Sobjects}{$\mathcal{O}$}
\newcommand{\Smetadata}{$\mathcal{S}$}
\newcommand{\token}[1]{\textless{}#1\textgreater{}}
\newcommand{\nibf}[1]{\noindent\textbf{#1}\ }

\newcommand{\propmodel}{\textit{UniQer}}
\newcommand{\propmodelL}{\textit{Unified Questioner Transformer}}

\newcommand{\oet}{\textit{OET}}
\newcommand{\qdt}{\textit{QDT}}
\newcommand{\otgt}{\textit{OTM}}

\newcommand{\sepmode}{}

\usepackage[pagebackref=true,breaklinks=true,colorlinks,bookmarks=false]{hyperref}

\arxivfinalcopy % *** Uncomment this line for the final submission

\ifarxivfinal\fi

\begin{document}

%%%%%%%%% TITLE
\title{Unified Questioner Transformer for Descriptive Question Generation\\in Goal-Oriented Visual Dialogue}

\author{Shoya Matsumori% \textsuperscript{\dag{}}
\thanks{Corresponding Author}
\thanks{Equally Contributed}
,
% {\tt\small shoya@ailab.ics.keio.ac.jp}
% For a paper whose authors are all at the same institution,
% omit the following lines up until the closing ``}''.
% Additional authors and addresses can be added with ``\and'',
% just like the second author.
% To save space, use either the email address or home page, not both
Kosuke Shingyouchi\textsuperscript{\dag{}},
Yuki Abe,\\
Yosuke Fukuchi,
Komei Sugiura,
Michita Imai\\ 
Keio University\\
{\tt\small shoya@ailab.ics.keio.ac.jp}
% 3--14--1 Hiyoshi, Kohoku Ward\\
%   Yokohama, Kanagawa 223--8522
}

\maketitle

%%%%%%%%% ABSTRACT
\begin{abstract}
Building an interactive artificial intelligence that can ask questions about
the real world is one of the biggest challenges for vision and language problems. 
In particular, goal-oriented visual dialogue, where the aim of the agent is
to seek information by asking questions during a turn-taking dialogue, has been gaining scholarly attention recently.
While several existing models based on the \gw{} dataset~\citep{de2017guesswhat} have been proposed,
the Questioner typically asks simple category-based questions or absolute spatial questions.
This might be problematic for complex scenes where the objects share attributes, 
or in cases where descriptive questions are required to distinguish objects.
In this paper, we propose a novel Questioner architecture, called \propmodelL{} (\propmodel{}),
for descriptive question generation with referring expressions.
In addition, we build a goal-oriented visual dialogue task called \task{}. 
It synthesizes complex scenes that require the Questioner to generate descriptive questions.
We train our model with two variants of \task{} datasets.
The results of the quantitative and qualitative evaluations show that \propmodel{} outperforms the baseline.
\end{abstract}

%%%%%%%%% BODY TEXT
\section{Introduction}\label{sr:intro}
% 1.1: Background
Information seeking through interaction is one of the most vital abilities for artificial intelligence.
This is particularly true in the human-agent interaction scenario~\cite{satake2009approach,kanda2017human}.
% What kind of cases that agents need to ask question in the interaction scenario?
For example, task-oriented agents need to understand what the users are thinking,
i.e., beliefs, preferences, and intentions, % preferences or desires?
in order to correctly interpret their instructions~\cite{iwahashi2010robots,zuo2010detecting}.
In most cases, such information is not provided prior to the interaction,
so the agents have to elicit it by asking questions on the fly.

\begin{figure}[tbp]
\centering
  \includegraphics[clip,width=\linewidth]{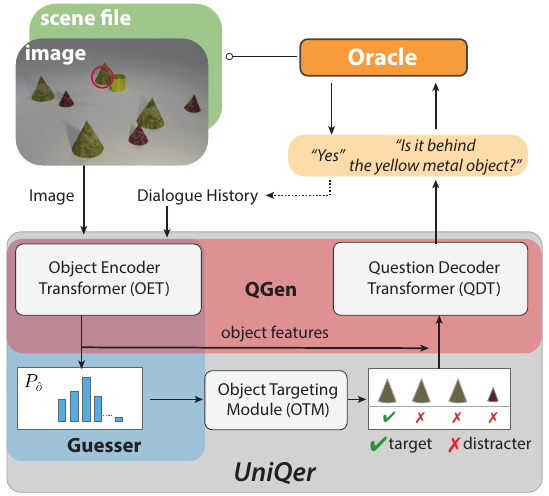}
  	\caption{
  	\propmodel{} and \task{} task.
      \propmodel{} unifies the Question Generator (QGen),
      which generates yes-no questions,
      and the Guesser, which guesses an \oracle{}'s reference object,
      into a single transformer encoder-decoder architecture.
      The Object Targeting Module is introduced to assign target objects
      which are to be set apart from distracter objects in the QGen.
    }
  	\label{fig:1}
  	\vspace{-5mm}
\end{figure}

% 1.2: Research Background
To build effective information-seeking agents,
several studies in the vision and language community have tackled
the goal-oriented visual dialogue task~\cite{de2017guesswhat,strub2017end}.
This task consists of two agents, called a \questioner{} and an \oracle{},
and the goal is to train the \questioner{} to guess the \oracle{}'s
reference object by asking yes-no questions during a turn-taking dialogue. 
The \oracle{} then needs to provide an answer given the question and the target object.

% 1.3 Our targets
In goal-oriented visual dialogue,
deciding which objects and how to address will depend on the complexity of the presented image.
For example, in simple scenes,
where each object has different attributes and thus is easy to distinguish,
the \questioner{} only needs to ask category-based questions such as
``\textit{Is it a car?}''.
In a more complex scene, on the other hand,
the \questioner{} needs to ask descriptive questions with referring expressions~\cite{dale1989cooking, winograd1972understanding}
to narrow down the candidates,
such as ``\textit{Is it behind a tree?}'' or ``\textit{Is it a red car?}''.
% when it only knows that the reference object is a car.
In this paper, we particularly focus on building an agent
that can facilitate such descriptive questions in goal-oriented visual dialogue.

% 1.3 Previous Studies and Issues
% Issuses within the previous studies
Several models based on \gw{}~\cite{de2017guesswhat} have been published, 
most of which leveraged reinforcement learning
aimed at maximizing the success reward by generating a word token
as an action~\cite{strub2017end,zhang2018goal,zhao2018improving,shekhar2018beyond,shukla2019should,abbasnejad2019s,pang2020visual,abbasnejad2020gold}.
% Several studies have been conducted to attack this problem, but still remain in marginal improvement.
However, the \questioner{}s in these models typically generate only simple
category-based questions, such as ``Is it a person?'' or ``Is it a computer?'',
or absolute spatial questions, such as ``Is it in front?'' or ``Is it on the left side of the image?'',
which is not effective when a large number of similar-looking candidates is presented in the same place.

% ¶ [Proposal] Interactive Visual Question Generation
In response to the above issues,
we propose \propmodelL{} (\propmodel{}) and a task called \task{}
for descriptive question generation in goal-oriented visual dialogue (Fig.~\ref{fig:1}).
In \propmodel{}, the question generator (QGen) and the Guesser are
unified into a single transformer architecture.
By utilizing such architecture, both the QGen and the Guesser can make use of the same object features,
and the Guesser can consider object relations more effectively thanks to the self-attention structure.
% We also adopt a target-wise object-question generation architecture.
We also introduce an object-targeting module,
inspired by the notion of target and distracter~\cite{dale1995computational},
which will decide on the objects that are to be addressed in the question.
With this structure, the QGen can focus on generating questions that will
address the target from other objects in a supervised manner.
% To build a \questioner{} that can ask descriptive question, a novel task called \task{} is proposed.
Our task, \task{}, includes complex scenes in which similar-looking %identical might be a word that fits here
objects are presented;
therefore, the \questioner{} needs to develop the ability to ask descriptive questions rather than simple ones.
The experimental results show that our model's performance exceeds that of the baseline
in terms of task success rate by around 20$\%$.
Besides, the extensive ablation studies conducted show the structural advantages of our model.

% ¶ [Contributions]
% Contributions
To summarize, our contributions are three-fold:
\vspace{-0.5em}
\begin{itemize}
 \setlength\itemsep{-0.4em}
  \item We proposed a novel unified transformer architecture for the \questioner{}.
  To the best of our knowledge, this is the first study that introduces
  a unified transformer architecture to goal-oriented visual dialogue.
\item To address the limitations of \gw{} dataset,
    we constructed a novel goal-oriented visual dialogue task, 
    namely \task{},
    which requires the \questioner{} to ask descriptive questions.
  \item We evaluated \propmodel{} with two variants of \task{} datasets and found that our model outperformed
    the baseline and was able to ask descriptive questions with referring expressions,
    given complex scenes where the objects were hard to distinguish.
\end{itemize}

\sepmode{}
\section{Related Works}\label{sr:related}
% \subsection{Visual Question Answering (VQA)}
\nibf{Visual Question Answering.}
Visual Question Answering (VQA)~\cite{antol2015vqa,johnson2017clevr,hudson2018compositional}
lies at the intersection of computer vision and natural language processing (NLP).
Compared with image captioning tasks~\cite{fang2015captions},
VQA requires a comprehensive understanding of the visual object elements and the relationships between them.
\citet{johnson2017clevr} proposed \clevr{}, a synthetic VQA dataset,
aiming to remove the difficulties in image recognition
and creates a balanced dataset that requires reasoning abilities without shortcuts.
Besides, extending VQA to dialogues has been challenged
in Visual Dialogue~\cite{visdial,schwartz2019factor}.

\nibf{Transformer Architecture.}
Transformer architecture~\cite{vaswani2017attention} was recently introduced in
NLP tasks showing the effectiveness of the self-attention structure in language modeling,
followed by large-scale pre-trained models such as BERT~\cite{devlin2018bert}
and its successive models~\cite{liu2019roberta, yang2019xlnet}.
Several studies have imported transformer architecture to the
aforementioned vision and language tasks,
such as Image BERT~\cite{qi2020imagebert}, Meshed Memory Transformer~\cite{cornia2020meshed},
and UNITER~\cite{chen2019uniter}.

\nibf{Referring Expression Generation.}
Referring expressions are language
constructions used to identify specific objects in
a scene~\cite{winograd1972understanding, dale1989cooking}.
These specific objects here are called the targets,
and a set of objects that are to be isolated from the target set
is called the distracter group~\cite{dale1995computational}.
Various datasets have been proposed recently
including both synthetic~\cite{liu2019clevr}
and natural image datasets~\cite{luo2017comprehension, yu2016modeling}
to generate and comprehend referring expressions.

\nibf{Goal-oriented Visual Dialogue.}
Our task is grounded on the goal-oriented visual dialogue framework.
This test-bed was first implemented in \gw{}~\cite{de2017guesswhat},
which is composed of 155K goal-oriented dialogues, collected via the Amazon Mechanical Turk,
and includes 822K questions, with a unique vocabulary size of 5K.
The images were borrowed from the \coco{} dataset and consist of up to 67k images and 134K objects.

% The explanations for models
The \gw{} task was originally designed for supervised learning,
but it was extended to fit into the reinforcement learning framework~\cite{strub2017end}.
Ongoing works promote various approaches to \gw{}.
One is the rewarding in reinforcement learning,
where \citet{zhang2017automatic} used two intermediate rewards,
while another, proposed by \citet{shukla2019should}, makes use of an informativeness reward based on regularized information gain.
\citet{zhao2018improving} improved the RL optimization by extending a policy gradient method
using a temperature for each action based on action frequencies.
Moreover, other studies focus on improving the model's architecture.
\citet{abbasnejad2019s} introduced a Bayesian approach to quantify the uncertainty in the model.
\citet{pang2020visual} proposed dialogue state tracking module to make use of the belief of the Guesser.
\citet{shukla2019should} proposed single visually grounded dialogue encoder
shared by both the Guesser and the QGen, trained with cooperative learning.
% built~\cite{zhang2018goal,zhao2018improving,shekhar2018beyond,shukla2019should,abbasnejad2019s,pang2020visual,abbasnejad2020gold}.

\nibf{Limitations.}
Among such models proposed on \gw{}, there are two major limitations.
(1)~First is the separated learning approach that previous methods adopt.
In most of the previous methods, a \questioner{} has two major components:
the QGen which generates questions based on the image presented and the current dialogue history
and the Guesser which guesses the target object of the \oracle{}.
Ideally, the object's features obtained during the training should be shared with both module,
but these two components are often learning separately.
Additionally, the Guesser in the previous studies
only looks at a single object at a time to determine the probability of it being the target object
and does not consider the relation between objects.
This will be problematic for processing questions that refer to the other objects, i.e. referring expressions.
(2)~Another problem comes from the fact that the QGen was too heavily burdened.
The QGen in the previous models needs to decide on which objects to refer to and how to refer to in the same architecture.
Coupled with the difficulties of tuning generator with reinforcement learning,
this will degrade the final performance of the task,
such as losing the lexical diversity of the questions as reported in \cite{shekhar2018beyond}.

% What will be the problems when it comes to generate questions with referring expressions.
While \gw{} has pioneered the frontier of the goal-oriented visual dialogue,
it has several issues yet to be resolved in order
to build a \questioner{} that can ask descriptive questions.
The major issue of the \gw{} dataset is the poor performance of the \oracle{}~\cite{de2017guesswhat}.
Since the \oracle{} is solving a 3-class classification problem---yes, no, and not applicable---,
this accuracy causes substantial errors in the \questioner{}’s learning process.
Another problem is that the \oracle{} generates answers
to the questions without having been given the visual features;
only the categorical information and spatial information of the objects are given.
This means that the \oracle{} is incapable of
understanding descriptive questions with referring expressions that include visual attributes.

\sepmode{}
\section{\task{} Specifications}\label{sr:task}
\nibf{Notation.}
\task{} is defined by the tuple $(\mathcal{I}, \mathcal{S}, \mathcal{O}, o^*, \mathcal{D})$,
where \Simage{} is the image, \Smetadata{} is the scene meta-data for the image,
\Sobjects{} is the set of objects appearing in the scene,
$o^*$ is the goal reference object of the \oracle{},
and \Sdialogue{} is the dialogue between the \oracle{} and the \questioner{}.

% Both image is shared by oracle and questioner
Formally, \Simage{} $\in\mathbb{R}^{H \times W}$ is the current observed image, with height $H$ and width $W$.
% Metadata
Corresponding to the image, the scene meta-data \Smetadata{} provides the information of
the scene, including the collections of the objects attributes
(shape, size, color, and material) and position on the ground-plane.
These are represented by a scene graph~\cite{johnson2015image} of \clevr{},
which provides a complete view of the environment.
% Target objects (the number of the objects)
The objects in the scene are represented by $\mathcal{O}={\{o_i\}}_{i=1}^{N}$,
where $o_i$ is the $i$-th object and $N$ is the number of the objects within the scene.
For each session, the goal object $o^* \in \mathcal{O}$ is arbitrarily chosen.
% Dialogue
The dialogue, where the \questioner{} seeks to identify the goal object $o^*$,
consists of question-answer pairs $\mathcal{D} = {(q^t, a^t)}_{t=1}^{T}$,
 $T$ denotes the number of such pairs.
Each question is composed of word tokens $q^t = {(w_\omega^t)}_{\omega=1}^{W_t}$,
that were sampled from the vocabulary $\mathcal{V}$.
% including both question tokens and special tokens.
Finally, the answer to the question is restricted to yes,
no or invalid question, that is, $a^t \in $~\{\token{yes}, \token{no}, \token{invalid}\}.

\nibf{Oracle.}
% What is the role of the oracle?
The central role of the \oracle{} is to provide an answer to each question generated by the \questioner{} based on the current scene.
Since its performance will directly affect the learning process of the \questioner{},
the \oracle{} needs to be perfect, as the name implies.

% restrict the language used in the questioner
To satisfy this demand, we built a robust \oracle{} function that meets the following two requirements.
% Complete understanding of the world
First, the \oracle{} needs to understand the scene completely to answer questions.
For this purpose, we take advantage of the fully structured environment of \clevr{}.
% Here, we introduced scene files in \clevr{} world.
Since the \clevr{} world is built on a structured ground-truth representation~\cite{johnson2015image},
the complete and exhaustive information about the image is available in the scene file.
We introduced such information to the \oracle{} so that it can
answer any kind of detailed expressions.
% We use this scene file to the \oracle{}
Second, the \oracle{} is required to completely understand what the \questioner{} says.
Here, we standardized the language used by the \questioner{} to a pseudo-language
that can be directly interpreted by the \oracle{}.
This language is an executable functional program;
given the scene-file, it yields the objects that match the query in a deterministic manner,
which enables the \oracle{} to deduce the answer to the question.
These settings will liberate the \oracle{} from having to interpret the question
based on the model obtained via learning, which is likely to produce some errors.

% What is the structure of the network?
\begin{figure*}[t] \begin{center}
  \includegraphics[clip,width=\linewidth]{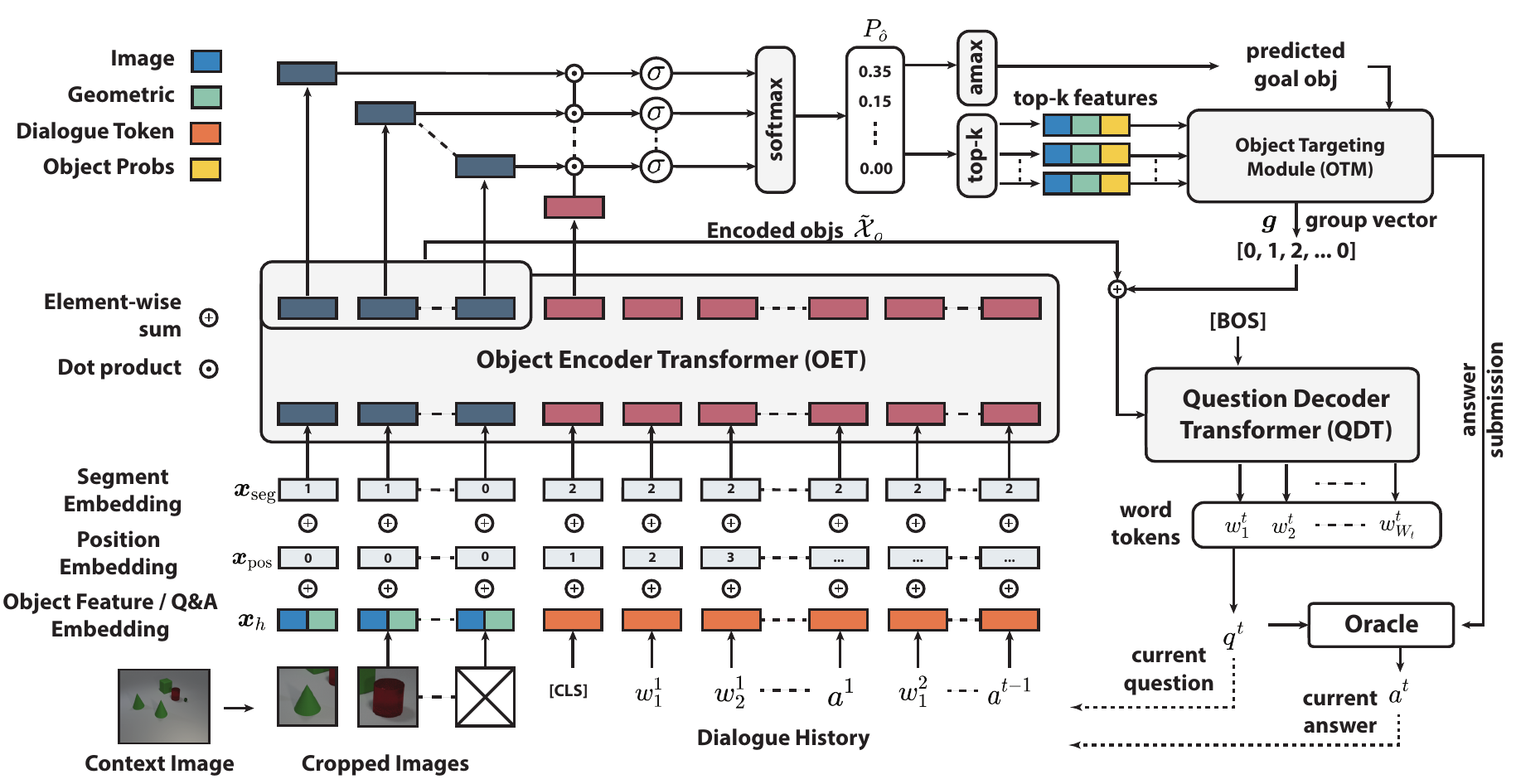}
  \caption{Network structure of \propmodel{}. 
  % The model consists of \oet{}, \otgt{}, and \qdt{} modules.
  Given the cropped image features, geometrical features, [CLS] token, and past question and answer tokens,
  with the segment and sequence position embedding injected, embedding vectors are fed into \oet{}.
   By comparing the encoded [CLS] token with the encoded objects and passing them to softmax, 
   goal object probability $P_{\hat{o}}$ is obtained.
   The top-k objects with high $P_{\hat{o}}$ are encoded and input to the \otgt{} module
   along with the image features and geometrical features, and each object is assigned a group ID.
   Finally, \qdt{} takes [BOS] as an input, generates the next question by taking the sum of
   the encoded objects and the group vector as memory, and the \oracle{} answers to the question.
  }\label{fig:model}
  \end{center}
\vspace{-5mm}
\end{figure*}

\nibf{Dataset Generation.}
The \task{} dataset consists of images including scene files and questions that are bound to images.
In the original \clevr{} dataset, the object attributes are
three object shapes, two absolute sizes, two materials, and eight colors.
Notably, the numbers of entities in each attribute in the \clevr{} dataset are not the same.
% As we tested in advance,
% \txmemo{Preliminary tests showed that}
This will likely cause undesirable shortcuts for the goal-oriented visual dialogue task;
for example, only asking questions about attributes with fewer entities, 
(e.g., sizes and materials) 
can result in the candidate objects being bisected.

To circumvent this issue, we prepared two new balanced datasets,
Ask3 and Ask4, which contain three and four entities for each attribute, respectively.
Both Ask3 and Ask4 datasets consist of  70K training, 7.5K validation, and 7.5K test images.
All images were generated by randomly sampling a scene graph and rendered by Blender~\cite{blender2016blender}.
We followed the original sampling procedure in \clevr{} except that for each generation step we randomly chose either to
copy the existing object or to create a new object by sampling attributes.
This enables us to intentionally place an identical object in the scene.

% What kind of questions?
As presented in~\cite{shekhar2018beyond}, the questions that appear
in goal-oriented visual dialogue can be roughly categorized into the following
question types and its combinations thereof:
entity, attribute, location, and action.
% Why we focus on attribute and location question
Among these, we focus on attribute questions and location questions with referring expressions.
The attribute question inquires about the attributes of the object
and the location questions ask about the spatial relations,
such as ``Is it the object in front of the purple cube?''.

% How are questions generated?
Questions are generated by sampling the question templates given the image scene graph.
Each question is validated to ensure it is related to the scene by checking
whether it is asking about something that exists in the scene.
Detailed statistics for the dataset are available in the appendix section.

\sepmode{}
\section{\propmodelL{}}\label{sr:model}
We propose \propmodelL{} (\propmodel{}) as a \questioner{} model for goal-oriented visual dialogue tasks.
Conceptually, \propmodel{} can be divided into the following three major components:
\begin{itemize}
    \setlength\itemsep{0em}
    \item Object Encoder Transformer (\oet{})
    \item Object Targeting Module (\otgt{})
    \item Question Decoder Transformer (\qdt{})
\end{itemize}
as shown in Fig.~\ref{fig:model}.
The \oet{} encodes objects and serves as a Guesser, which infers the goal object $o^*$.
The \otgt{} is trained to learn which object should be addressed in the next question
based on the \oet{}'s current inference. % with reinforcement learning.
The \qdt{} decodes object features and generates questions,
by addressing the object to set apart from the other objects
that have different group IDs from the \otgt{}.
When it is confident enough, the \otgt{} decides to submit its answer to the oracle.
Both the \oet{} and \qdt{} are trained jointly in a supervised manner,
and thus they can benefit from each other.
The \oet{} is trained with reinforcement learning.

\subsection{Embedding Preparation}\label{sec:embd}
% All inputs to the \propmodel{} are first encoded in the embedding layer.
The input of the \propmodel{} is the cropped object images and dialogue history.
The embedding layer prepares the feature vectors for both inputs by
\begin{align}
    \bm{x}_{e}(i) = \bm{x}_h(i) + \bm{x}_{\text{seg}}(i) + \bm{x}_{\text{pos}}(i),
\end{align}
where $i$ denotes the index of the object,
$\bm{x}_{h}(i)$ is either the object feature embedding $\bm{x}_{v}(i)$
or the dialog token embedding $\bm{x}_{l}(i)$,
$\bm{x}_{\text{seg}}(i)$ is a segment embedding, and $\bm{x}_{\text{pos}}(i)$ is a sequence position embedding.

\nibf{Object Feature Embedding.}
Each cropped image corresponding to the objects in the scenes
is first processed to $\bm{o}_{v}(i)$ using an arbitrary image feature extractor.
As in~\cite{yu2017joint}, two types of 5D geometric feature 
for each object are introduced as well: a source geometric feature represented by
$\bm{o}_{sg}(i) = [
\frac{x_{ci}}{W},
\frac{y_{ci}}{H},
\frac{w_{bi}}{W},
\frac{h_{bi}}{H}, 
\frac{w_{bi}h_{bi}}{WH}
]$,
and relative geometric feature
$\bm{o}_{rg}(i, j) = [
\frac{x_{cj}}{w_{bi}},
\frac{y_{cj}}{h_{bi}},
\frac{w_{bj}}{w_{bi}},
\frac{h_{bj}}{h_{bi}}, 
\frac{w_{bj}h_{bj}}{w_{bi}h_{bi}}
]_{j \in \mathcal{N}, j\neq{}i},
$
where $(x_{ci}, y_{ci})$ and $(w_{bi}, h_{bi})$
denote the center coordinates and the width and height of the bounding box for the object $o_i$, respectively, $(W, H)$ is the scene image size,
and $\mathcal{N}$ is the set of object indices in the scene.
Thus, we define the geometrical feature as follows: 
\begin{align}
\bm{o}_{g}^{\Phi}(i) = [\bm{o}_{sg}(i), \{\bm{o}_{rg}(i, j)|j\in{\Phi},j\neq{}i\}],
\end{align}
where $\Phi$ is a set of object indices.
Then the image feature and the geometrical feature are concatenated and fed into the linear 
transformation function $\mathcal{F}_{v}$ with ReLU activation as follows:
$\bm{x}_{v}(i) = \mathcal{F}_{v}([\bm{o}_{v}(i), \bm{o}_{g}^{\mathcal{N}}(i)])$.

\nibf{Dialogue Token Embedding.}
Questions and answers in the current dialogue history
$\mathcal{D}^{t} = {(q^\tau, a^\tau)}_{\tau = 1}^{t}$ are transformed into
dialogue token embedding $\bm{x}_{l}^{t}$ with a one-hot embedding layer. 
Additionally, the classification token ([CLS])~\cite{devlin2018bert},
is introduced at the very front of the dialogue token.

\nibf{Segment embedding and sequence position embedding.}
% Why segment embedding was introduced?
Segment embedding $\bm{x}_{\text{seg}}$ is introduced to distinguish the modality of the input.
There are three tokens that represent the cropped object image input, 
dialogue token input, and mask~(padding) token input.
% Why sequence position embedding?
Sequence position embedding $\bm{x}_{\text{pos}}$ is introduced to indicate the position of the token in the sequence.
We use the same value for all object embeddings since there is no numerical order in objects,
% and geometric information has already been added to the embeddings,
while for tokens in the dialogue history, we add the position ID incrementally.

\subsection{Object Encoder Transformer}\label{sec:oenctrm}
% The purpose of the encoder transformer.
Given the input embeddings $\bm{x}_{e}$, the \oet{} plays two important roles;
it extracts object features that are then used in the \qdt{} and the \otgt{}
and it predicts a goal object.
The \oet{} is built on a transformer encoder architecture~\cite{vaswani2017attention}.

The output feature %I don't think it's necessary to write feature
of the \oet{} is written as
$\tilde{\mathcal{X}}=\{\tilde{\mathcal{X}}_{o}, \tilde{X}_{[\rm{CLS}]}, \tilde{\mathcal{X}}_{l}\}$,
where $\tilde{\mathcal{X}}_{o} = (\tilde{X}_{o}^{1}, \ldots, \tilde{X}_{o}^{N_{\mathrm{max}}})$
is a set of object embeddings with the maximum number of objects $N_{\mathrm{max}}$,
$\tilde{X}_{[\rm{CLS}]}$ is the embeddings of the [CLS] token, and
$\tilde{\mathcal{X}}_{l} = (\tilde{X}_{l}^1, \ldots, \tilde{X}_{l}^{\Omega})$ is a set of dialogue question and answer token embeddings with its current length $\Omega$.
From the encoder output $\tilde{\mathcal{X}}$,
only the object embeddings $\tilde{\mathcal{X}}_{o}$ are passed to the \qdt{}.

As for the goal object prediction,
the model needs to find the objects that match the current dialogue history $\mathcal{D}^{t}$.
First, the inner products of linear-transformed object embeddings $\tilde{\mathcal{X}}_o$
and [CLS] token embeddings $\tilde{X}_{[\text{CLS}]}$ are computed as
\begin{align}
    \sigma_{o} = [\mathrm{sigmoid}(
    \mathcal{F}_{o}(\tilde{X}_{o}^{i}) \cdot \mathcal{F}_{c}(\tilde{X}_{[\text{CLS}]})
    )]_{i \in \mathcal{N}},
    \label{eq_sigobj}
\end{align}
where $\mathcal{F}_{o}$ and $\mathcal{F}_{c}$ are the linear transformation functions.
Then, we apply the $\mathrm{softmax}$ function to acquire the goal object probability
\begin{align}
  P_{\hat{o}} = \mathrm{softmax}(\sigma_{o})\label{eq_prob}.
\end{align}
Similar to the role of the dialogue state encoder in previous studies,
$\tilde{X}_{[\mathrm{CLS}]}$ is expected to capture the summary of the dialogue that is comparable to the object features in order to produce the similarity.

\subsection{Object Targeting Module}\label{sec:otrg}
The \otgt{} has two roles;
it determines which objects should be addressed by \qdt{}
and
it decides when to submit the answer, the \oet{}'s prediction, to the \oracle{}.

% A discriptions for object targeting (3groups)
The \otgt{} determines which objects should be addressed in
the question by assigning any of the following three types of property to each object:
a target object property for the object to be addressed,
a distracter object property for an object that will not be addressed but is distinguishable from the target objects,
and a masked object property for objects that do not exist or will be ignored.

% An input and an output of the OTM.
The \otgt{} only pays attention to top-k high-scored candidate objects calculated
from the object probability $P_{\hat{o}}$, and ignores the others.
More specifically, the input of the \otgt{} is the set of top-k feature vectors
$\{\bm{x}_{k}(i)\}_{i\in\mathcal{K}}$, 
where $\mathcal{K}$ is the set of top-k object indices and
$\bm{x}_{k}(i) = [\mathcal{F}_{A}(\bm{o}_{v}(i)), \mathcal{F}_{B}(\bm{o}_{g}^{\mathcal{K}}(i)), \mathcal{F}_{C}({P}_{\hat{o}}(i))]$,
where $\mathcal{F}_{A}, \mathcal{F}_{B}$, and $\mathcal{F}_{C}$ are different linear layers.
  Given such an input, the \otgt{} will decide how to assign these properties to each top-k objects
  by producing a targeting action $\hat{g}_k$ as
\begin{align}
  \hat{g}_k \thicksim \mathcal{F}_{\mathrm{RL}}
  ( \{ \bm{x}_{k}(i) \}_{i\in\mathcal{K}} )\label{eq_gk},
\end{align}
    where $\hat{g}_k$ is an integer ranging from zero to $3^{k}-1$,
    which corresponds to the number of combinations to respectively
    allot the three property groups to $k$ objects.
$\mathcal{F}_{\mathrm{RL}}$ is a parameterized function trained with RL defined as
\begin{align}
  \mathcal{F}_{\mathrm{RL}} &= \mathrm{softmax}
  (\mathcal{F}_{l}([\mathcal{F}_{\text{GRU}}(\{\bm{x}_{k}(i)\}_{i\in{\mathcal{K}}})]))\label{eq_rl},
\end{align}
where $\mathcal{F}_{\text{GRU}}$ is the two-layered bi-directional GRU
to encode input vectors and 
$\mathcal{F}_{l}$ is the linear transformation function with the ReLU activation.

  To make the \otgt{}'s output compatible with the \qdt{}'s input,
  the targeting action $\hat{g}_k$ is then reformatted with base-10 to base-3 conversion
  to a top-k group vector $\bm{g}_k~\in~\mathbb{R}^{k \times 3}$,
  whose elements each represent the group ID --zero, one, or two-- assigned to an object.
  Finally, a group vector $\bm{g}~\in~\mathbb{R}^{N_\mathrm{max} \times 3}$ is
  obtained by filling the masked property group ID to the index for the ignored objects on top-k.

  We treat the cases
  $\hat{g}_k = 0$ and $\hat{g}_k = 3^{k}-1$
  as a submission action,
  which submits the current answer $\hat{o} = \mathrm{argmax(P_{\hat{o}})}$ to the oracle.
  Examples of the OTM are demonstrated in the appendix section.

\subsection{Question Decoder Transformer}\label{sec:qdectrm}
The role of the \qdt{} is to generate a question that
will distinguish the target objects from the distracter objects
as specified by the group vector $\bm{g}$ produced by the \otgt{}.
The \qdt{} consists of a standard transformer decoder~\cite{vaswani2017attention}.
The input of the \qdt{} is the beginning of sentence token ([BOS])
and as a memory we put the element-wise sum of the object embeddings $\tilde{\mathcal{X}}_o$
and the embeddings of group vector $\bm{g}$, described as:
\begin{equation}
  \mathcal{M} = \{
  \tilde{X}_{o}^{(i)} + \mathcal{F}_{s}(\bm{g}^{(i)})
  \}_{i\in\mathcal{N}}, \label{eq_qgen_input}
\end{equation}
where $\mathcal{F}_{s}$ is an embedding function 
and $\bm{g}^{(i)}$ is the i-th element of $\bm{g}$,
which represents the property group of the object $o_i$.
Given [BOS] token and $\mathcal{M}$, the \qdt{} generates a tokenized word in an auto-regressive manner.

\sepmode{}
\section{Training Process}\label{sr:training}
\subsection{Supervised Learning.}\label{ssr:train_super}
Both the \oet{} and the \qdt{} are jointly trained with supervised learning
to predict the goal object and generate questions.
For this purpose,
the overall loss function can be expressed as the sum of the object prediction loss and the target-wise question generation loss, as
$L = \alpha L_{\mathrm{pred}} + L_{\mathrm{gen}}$, 
where $\alpha$ is a constant that modifies the ratio between the two loss functions.
% and is responsible for the trade-off between \oet{} and \qdt{} performance.

\nibf{Object Prediction Loss.}
% multi-label classification problem
To predict the goal object,
we can simply classify each object as to whether or not it is a candidate at the end of each question and answer pair. By applying the $\mathrm{softmax}$ function during reinforcement learning,
we can acquire the goal object probability. Therefore, the loss for candidate object prediction is formalized as a multi-label classification problem,
where the binary cross-entropy is applied to
the sigmoid-activated result in Eq.~(\ref{eq_sigobj}) as follows:
\begin{equation}
  L_{\mathrm{pred}} = \sum_{t=1}^{T} -y_o^t \log(\sigma_{o}^{t})
  -(1-y_{o}^{t})\log(1-\sigma_{o}^{t}),\label{eq_oploss}
\end{equation}
where $T$ is the number of questions in a dialogue,
$y_o^t$ represents the ground-truth binary labels for the object in the $t$-th question.

\nibf{Target-Wise Question Generation Loss.}
Question generation loss is defined as the negative log-likelihood, as
\begin{equation}
  L_{\mathrm{gen}} = -\sum_{t=1}^{T}\sum_{l=1}^{W_t}
  \log p(w_{l+1}^t|w_{l}^t,\ldots, w_{2}^t, w_{1}^t, \mathcal{M}),
\end{equation}
where $W_t$ is the number of tokens included in the $t$-th question and $\mathcal{M}$
is expressed as
$\{\tilde{X}_{o}^{(i)} + \mathcal{F}_{s}(\bm{g}^{(i)})\}_{i\in\mathcal{N}}$.
During supervised learning,
the $\bm{g}$ is computed from the ground-truth question.
Here, the target group is assigned to the object that matches with the question when the answer to it is true, otherwise the distracter group is assigned.
Note that, $N_\mathrm{{max}} - k$ objects are assigned to a masked group in order to reproduce the reinforcement training conditions,
where $k$ is a pre-defined constant for top-k objects.

\subsection{Reinforcement Learning}\label{ssr:train_rl}
The \otgt{} is trained with reinforcement learning to
generate the group vector $\bm{g}$ and decide when to submit its answer.
During RL, the \oet{} and the \qdt{} are frozen.

% \nibf{Formalization.}
% section: reinforcement learning
We formalize \task{} as an MDP problem given the tuple $\left(S, A, P, R, \gamma \right)$,
where $S$ is the set of states, $A$ is the finite set of actions, $P$ is the state transition function,
$R$ is the reward function, and $\gamma$ is the discount factor.
% action space
% introduction
We define each state, action, and reward on the timestep $t$.
% actions
The set of actions $A$ corresponds to $\hat{g}_k$
produced by an action function $F_{\mathrm{RL}}$ defined in Eqs.~(\ref{eq_gk},~\ref{eq_rl}).
% states
The states of the game are defined as $S_{t} = (\mathcal{I}, \{(q^{\tau}, a^{\tau})\}_{1:t-1})$.
% special case for <EOD>
Agent choose actions until the number of questions reaches the pre-defined limit question count, $T$.
Among the $3^k$ actions, we treat $A_{t} = 0$ or $A_{t} = 3^k - 1$ as end of dialogue~(EOD) cases.
If one of these actions is selected, the dialogue is considered finished and 
the object with the highest probability $P_{\hat{o}}$ at that time is submitted to the \oracle{} as the final prediction.
The \oracle{} compares the \questioner{}'s prediction with the ground truth $o^*$
and returns the reward.
The model is trained with policy gradient optimization using REINFORCE algorithm~\cite{sutton2000policy}.

\sepmode{}
\section{Experiments}\label{sr:exp}
\subsection{Settings}\label{ssr:settings}
\nibf{Datasets.}
We trained and evaluated our model with two different datasets: CLEVR Ask3 and CLEVR Ask4.
\\
\nibf{Baseline model.}
We compared the proposed model with a model proposed in \cite{strub2017end}.
This baseline model consists of a Guesser-QGen
architecture connected with a dialogue state encoder,
where the Guesser is a multi-label classifier with a single weight-shared MLP and the QGen is a recurrent neural network.
As in the previous studies, the QGen and Guesser are pre-trained separately
and then tuned with reinforcement learning using policy gradient optimization.
Details of the baseline model can be found in the appendix.

\nibf{Implementation details.}
Supervised learning was early-stopped with 50-epoch patience
using the AdaBelief optimizer~\cite{zhuang2020adabelief}
with the learning rate 1e-4, and 20 epochs for the warmup, and a batch size of 1024. 
As for the reinforcement learning,
all experiments were trained for 150 epochs using the Adam optimizer~\cite{kingma2014adam}
with the learning rate 5e-4, and a batch size of 1024. 
Additional details are available in the appendix.

\subsection{Supervised Learning}\label{ssr:exp_super}
\nibf{Metrics.}
In supervised learning, we evaluated the \oet{} and the \qdt{} with three metrics:
F1 score, perfect address ratio, and correct address ratio.

F1 score was used to evaluate if the \oet{} can find out
candidate objects that match a dialogue $\mathcal{D}$.
The F1 score was computed from predicted candidate objects $\widehat{\mathcal{O}}_{\mathcal{D}}$
and  ground-truth candidate objects $\mathcal{O}_{\mathcal{D}}$.
We obtained $\widehat{\mathcal{O}}$ from the result of $\sigma_i$ in Eq.~(\ref{eq_sigobj}) as
$\widehat{\mathcal{O}} = \{o_i~|~i \in \mathcal{N} \wedge 0.5 < \sigma_i\}$.

% Eval for: question decoder transformer
Perfect address ratio and correct address ratio were used to evaluate
the \qdt{}'s ability to generate questions that will distinguish the target objects from
the distracter objects as instructed by a group vector $\bm{g}$.
The group vector $\bm{g}$ groups objects $\mathcal{O}$ into three groups:
the target object group $\mathcal{O}_{\mathrm{t}}$,
the distracter object group $\mathcal{O}_{\mathrm{d}}$,
and the masked object group $\mathcal{O}_{\mathrm{m}}$.
If the generated question succeeds to distinguish $\mathcal{O}_{\mathrm{t}}$
from $\mathcal{O}_\mathrm{d}$, it is perfect.
If it succeeds to do so but mixes $\mathcal{O}_{\mathrm{m}}$ with $\mathcal{O}_\mathrm{t}$,
it is only deemed correct.
The detailed definition of these metrics is available in the appendix section.

\nibf{Results.}
The results of supervised learning are summarized in Tab.~\ref{tab:supervised}. 
For both Ask3 and Ask4 datasets, the F1 scores yield a near-perfect 0.994.
The question generation achieved a fairly high probability,
almost 87\%, of generating correct questions in Ask3 and
even in Ask4, which has an increased dataset complexity,
as it is still able to generate nearly 70\% of the questions correctly.
The scores on perfect address were around 58\% and 43\% for Ask3 and Ask4 respectively. 
Although they were lower then the scores of correct address,
the model still shows the capability of addressing the target objects
considering the irrelevant masked objects.
\begin{table}[t]
\centering
\small
\begin{tabularx}{.47\textwidth}{@{}p{2.0cm}ccc@{}}
\toprule
Model         & F1 score$\uparrow$ & Perfect Addr$\uparrow$ & Correct Addr$\uparrow$ \\ \midrule
UniQer (Ask3) & $0.994$  & $57.67$       & $86.91$                           \\
UniQer (Ask4) & $0.994$  & $43.20$       & $69.79$                           \\ \bottomrule
\end{tabularx}
\caption{
The results for supervised training for both Ask3 and Ask4 datasets.
F1 score measures the \oet{}'s ability to find out the object candidates given a dialogue.
Perfect Addr and Correct Addr stand for ``perfect address ratio'' and ``correct address ratio'', respectively, which measures the \qdt{}'s ability to generate a question
that correctly addresses the target objects instructed by a group vector $\bm{g}$.
}\label{tab:supervised}
\vspace{-1em}
\end{table}

\subsection{Reinforcement Learning}\label{ssr:exp_rl}
\nibf{Metrics and Conditions.}
Following~\citet{strub2017end},
we used the task success ratio,
defined as the rate of correct predictions submitted by a questioner,
as our primary metric.
% What is the test set? -> New Image and New Object
The training was conducted in two different settings: new image and new object.
In the new image setting, both the image and the goal object
are completely new and have not been presented before. 
In the new object setting,
the goal object is new but the image has already been presented in the training. 
We conducted five experimental runs across different seeds. % thanks

\nibf{Reward function.}
% How the reward is given?
The basic reward for reinforcement learning is the zero-one task success reward
% $r(S_t, A_t)$
$r_c$,
similar to~\citet{strub2017end},
which is given when the stop-action is produced and the predicted goal object is correct.
Note well that submitting the answer at the first action step is treated as invalid
and the success reward will not be given, so as to prevent the random predictions without asking questions.
We also introduced a turn discount factor $r_{d}$,
similar to the goal-achieved reward proposed by~\citet{zhang2018goal},
which will give a discount to the success reward
depending on the number of questions asked to reach the answer.

\nibf{Ablation Studies.}
Ablation studies were conducted to determine the effectiveness of \propmodel{}.
We performed testing using the following three ablation models:
\begin{itemize}
    \setlength\itemsep{-0.5em}
    \item \textbf{Vanilla:} The simplest model, which utilizes bi-directional GRU as the object encoder and LSTM as the question decoder. The Guesser architecture, a MLP module, is trained with the LSTM dialogue history encoder. However, it is trained separately from the object encoder, the same as the baseline model.
    \item \textbf{Not Unified MLP Guesser:}
    The model that substitutes the object encoder and
    the question decoder in the Vanilla model with a standard transformer.
    Note that the object encoder in this model does not include the Guesser architecture;
    as an alternative, the Guesser is implemented with a multi-layer perceptron.
    \setlength\itemsep{-0.5em}
    \item \textbf{Not Unified:} The model that substitutes the Guesser module with the transformer.
    This model is the disassembled version of \propmodel{}, whose QGen and Guesser modules are built on a single transformer encoder-decoder architecture.
\end{itemize}
\begin{table}[t]
\footnotesize

\begin{centering}
  \begin{tabularx}{.47\textwidth}{@{}p{1.3cm}p{1.3cm}p{1.3cm}p{1.3cm}p{1.3cm}@{}}
    \toprule
%               & \multicolumn{3}{c}{Task Success (\%)} \\ \midrule
    \multicolumn{1}{c}{} & \multicolumn{2}{c}{Ask3}                 & \multicolumn{2}{c}{Ask4}                                                      \\ \midrule
    Model                & New Img$\uparrow$ & New Obj$\uparrow$ & \multicolumn{1}{c}{New Img$\uparrow$} & \multicolumn{1}{c}{New Obj$\uparrow$} \\ \midrule
    Baseline  & $60.00_{\pm 6.35}$       & $59.60_{\pm 6.87}$      & $64.75_{\pm 0.82} $      & $64.21_{\pm 0.34}$      \\
    Ours (v)   & $72.98_{\pm 3.13} $      & $72.88_{\pm 3.47}$      & $67.38_{\pm 4.18}$       & $67.01_{\pm 4.34}$     \\
    Ours (num) & $69.43_{\pm 2.75} $      & $69.50_{\pm 2.99}$      & $72.89_{\pm 5.95} $      & $72.35_{\pm 5.94}$      \\
    Ours (nu)  & $50.61_{\pm 6.51} $      & $50.37_{\pm 6.02}$      & $65.15_{\pm 3.33} $      & $64.25_{\pm 3.01}$      \\
    \textbf{Ours (full)}& $\bm{84.10}_{\pm 4.41}$  & $\bm{83.96}_{\pm 4.70}$ &  $\bm{83.47}_{ \pm 1.25 } $ & $\bm{83.81}_{ \pm 0.94 }$\\
    \bottomrule
  \end{tabularx}
\end{centering}

% \captionsetup{width=.40\textwidth}
\caption{
Quantitative results on comparison and ablation study.
The average and standard deviation of five runs for the task success ratio are shown.
The bold numbers represent the best performance.
The model ``Ours (full)'' represents our proposed UniQer
and the other ``Ours'' models are ablated UniQer models;
``Ours~(v)'' is the \textit{Vanilla} model,
``Ours~(num)'' is the \textit{Not Unified MLP Guesser} model,
and ``Ours~(nu)'' is the \textit{Not Unified} model.
}\label{tab:results_rl}
\vspace{-1em}
\end{table}
\begin{table}[t]
\footnotesize
\begin{centering}
  \begin{tabularx}{.47\textwidth}{@{}p{1.29cm}p{1.3cm}p{1.3cm}p{1.3cm}p{1.3cm}@{}}
    \toprule
%                & \multicolumn{3}{c}{Task Success (\%)} \\ \midrule
    \multicolumn{1}{c}{} & \multicolumn{2}{c}{Ask3}                 & \multicolumn{2}{c}{Ask4}                                                      \\ \midrule
    Model                & New Img$\uparrow$ & New Obj$\uparrow$ & \multicolumn{1}{l}{New Img$\uparrow$} & \multicolumn{1}{l}{New Obj$\uparrow$} \\ \midrule
    Ours (rand) & $1.73_{\pm 0.12}$  & $1.64_{\pm{0.21}}$ & $1.61_{\pm 0.07}$  & $1.72_{\pm{0.10}}$ \\
    Ours (fs) & $68.78_{\pm 0.27}$ & $68.49_{\pm 0.38}$ & $68.99_{\pm0.37}$ & $69.73_{\pm0.45}$ \\  % self attention
    % Ours (SA) & $84.23_{\pm 3.93}$ & $84.04_{\pm 3.93}$ & $83.80_{\pm 0.83}$ & $83.43_{\pm 0.83}$ \\  \midrule % self attention 
    \midrule
    Ours (k=4) & 
    $81.51_{ \pm 3.72 } $ & $82.27_{ \pm 3.77 }$ & $\bm{83.47}_{ \pm 1.25 } $ & $\bm{83.81}_{ \pm 0.94 }$ \\
    Ours (k=5) & $\bm{84.10}_{\pm 4.41}$    & $\bm{83.96}_{\pm 4.70}$ & $81.20_{\pm 4.37} $ & $80.50_{\pm4.86}$  \\
    Ours (k=6) &
    $83.20_{ \pm 3.37 } $ & $83.82_{ \pm 3.27 }$ & $80.84_{ \pm 4.77 } $ & $81.18_{ \pm 4.79 }$ \\
    Ours (k=7) &
    $82.66_{ \pm 3.44 } $ & $83.39_{ \pm 3.12 }$ & $76.47_{ \pm 4.90 } $ & $76.63_{ \pm 4.57 }$ \\
    \bottomrule
  \end{tabularx}
  \caption{
    Ablated results on Object Targeting Module (OTM).
    We conducted ablation studies by changing the number of top-k
    and substituting the OTM with a random action module.
    The ``Ours (rand)'' is the random condition and
    the ``Ours (fs)'' is the force-stop random condition,
    which forcefully submits the prediction at the end of the dialogue.
}\label{tab:results_topk}
\end{centering}
\vspace{-1em}
\end{table}

% Design choices for OTM
Additionally, we conducted an extensive ablation to gain an understanding of the OTM.
In the ablation, we substituted the OTM with a random action model,
which chooses an action $A_{t}$ randomly on every step.
Besides, we tested this random model with force-stop condition,
where the model automatically submits the answer at the end of a dialogue,
which means the submission action is not required.
We also investigated the hyper-parameter settings for top-k.

\nibf{Results.}
The results on our full model are presented in Tab.~\ref{tab:results_rl}. 
In the table, the average and the standard deviation of task success ratio is shown.
The results demonstrate that \propmodel{} outperformed the baseline
with a large magnitude in both datasets and conditions,
showing \propmodel{}'s ability to discover a goal object
in the task which requires descriptive question generation.
In the new image setting,
the task success rate of the baseline model was 60.00\% and
64.75\% for Ask3 and Ask4 datasets, respectively,
while \propmodel{} achieved 84.10\% and 81.20\%.
% Why Ask3 < Ask4 in the baseline?
Surprisingly, the results with the new object setting was
on par with that of the new image setting,
which is likely due to the fact that their objects share the same attributes.

The ablated results are also in Tab.~\ref{tab:results_rl}.
As shown in the table, all of the ablated conditions
drop performances compared with \propmodel{},
showing the advantage of our architectural design.
Notably, the \textit{Not Unified} model significantly drops the performance among the others in both datasets.
This indicates the effectiveness of unifying the Guesser and the QGen.

The ablated results on the OTM module are shown in Tab.~\ref{tab:results_topk}.
The random condition scored the lowest as we expected,
while the force-stop condition performs much better than it.
This is because in the random condition, a chance of choosing a submission action is quite low.
The results indicate that learning when to submit the answer is vital function in the OTM.
We also found that, in our conditions, increasing $k$ does not always improve the results.

\nibf{Qualitative Results.}
\begin{figure}[tbp] \begin{center}
  \includegraphics[clip,width=\linewidth]{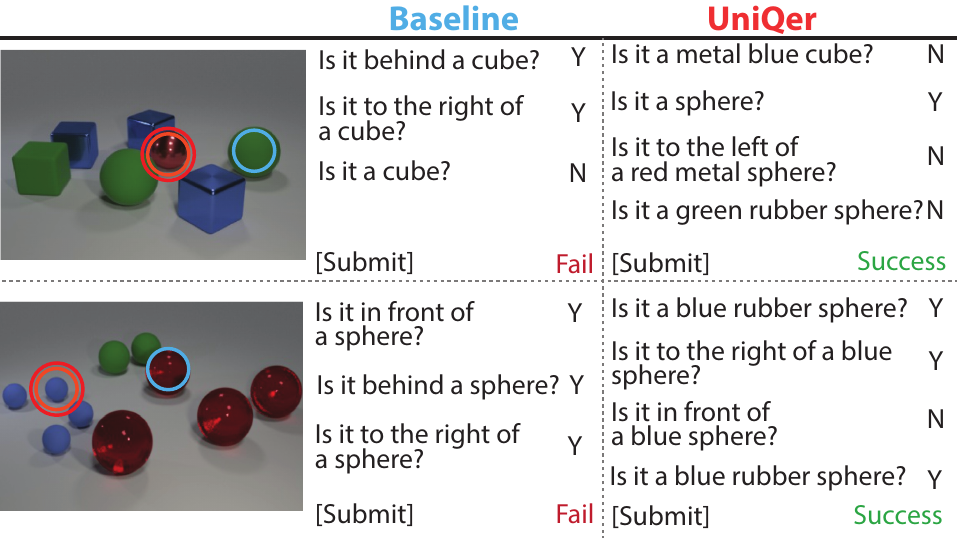}
  	\caption{
  	  Samples of generated dialogues converted to natural language.
      The red circles indicate the goal objects, the orange ones the \propmodel{} predictions, and the blue ones the baseline model predictions.
    }
    \label{fig:results}
  \end{center}
\vspace{-2em}
\end{figure}
We also inspected some of the dialogue samples and found
that \propmodel{} was actually capable of 
effectively generating descriptive questions to find out a goal object.
Fig.~\ref{fig:results} shows the representative dialogue examples for the test set images. 
In the upper example, the target object was the red metallic sphere.
The baseline generated questions regarding ``a cube'',
however none of the referring expressions were used
for narrowing down a referenced object among the cubes.
Because there are multiple cubes in the image,
this question is considered to be a non-informative question.
On the other hand, our method generated expressions such as ``a metal blue cube'' and ``a red metal sphere''. These questions are appropriate to distinguish the goal object among others, and considered to be informative.

In the lower example, the goal object was a small blue sphere which is located in a group of similar objects. The baseline generated referring expressions such as ``in front of a sphere'' and ``behind a sphere''. This is because all of the objects in the image were spheres and this is insufficient to disambiguate the goal object.
On the other hand, our method generated referring expressions such as ``right of a blue sphere'' and ``in front of a blue sphere.'' 
Although these expressions are not very human-friendly, they are sufficient and short enough to specify the relative relationships among the blue spheres.

\sepmode{}
\section{Conclusion}\label{sr:con}
In this research, we presented \propmodel{},
a novel \questioner{} architecture for
descriptive question generation with referring expressions in
goal-oriented visual dialogue.
Experimental results demonstrated that our model surpasses
the baseline on the novel \task{} datasets,
which require descriptive question generation.
We also validated the components of our model
with ablation studies, showing the structural advantages of our model.
Finally, we investigated the generated samples qualitatively
and found that our model successfully generated descriptive questions
and discovered the goal objects.

% TODO: remove for blinded review
%\subsection*{Acknowledgements}
%This work was supported by JST CREST Grant Number JPMJCR19A1, Japan.
% % % % % % %The experiments were partially conducted on AI Bridging Cloud Infrastructure (ABCI) \txtodo{do not use!}
%provided by National Institute of Advanced Industrial Science and Technology (AIST).
%The training progress was tracked and the reports were created with Weight \& Biases.
%Finally, we thank David Felices for helpful comments and discussion.

\clearpage
{
\small
\bibliographystyle{abbrv}
\bibliography{reference}
}
\end{document}